\documentclass[11pt,a4paper]{article}
\usepackage[hyperref]{acl2019}
\usepackage{times}
\usepackage{latexsym}
\usepackage{times}
\usepackage{epsfig}
\usepackage{graphicx}
\usepackage{amsmath}
\usepackage{amssymb}
\usepackage{url}
\usepackage{cuted,booktabs}
\usepackage{url}

\aclfinalcopy 

\title{Generating Question Relevant Captions to Aid Visual Question Answering}

\author{Jialin Wu, Zeyuan Hu and Raymond J. Mooney \\
  Department of Computer Science \\
  University of Texas at Austin \\
  \texttt{\{jialinwu, iamzeyuanhu, mooney\}@cs.utexas.edu}}

\date{}

\begin{document}
\maketitle
\begin{abstract}
Visual question answering (VQA) and image captioning require a shared body of general knowledge connecting language and vision. We present a novel approach to improve VQA performance that exploits this connection by jointly generating captions that are targeted to help answer a specific visual question. The model is trained using an existing caption dataset by automatically determining question-relevant captions using an online gradient-based method. Experimental results on the VQA v2 challenge demonstrates that our approach obtains state-of-the-art VQA performance ($e.g. $ 68.4\% on the Test-standard set using a single model) by simultaneously generating question-relevant captions.
\end{abstract}

\section{Introduction}
In recent years, visual question answering (VQA) \cite{antol2015vqa} and image captioning \cite{donahue2015long,rennie2017self} have been widely studied in both the computer vision and NLP communities. Most recent VQA research \cite{lu2017knowing,pedersoli2017areas,anderson2017bottom,lu2018r} concentrates on directly utilizing visual input features including detected objects, attributes, and relations between pairs of objects. 

\begin{figure}[!t]
\centering
\includegraphics[width=\linewidth,trim={0.5cm 1cm 16cm 0cm},clip]{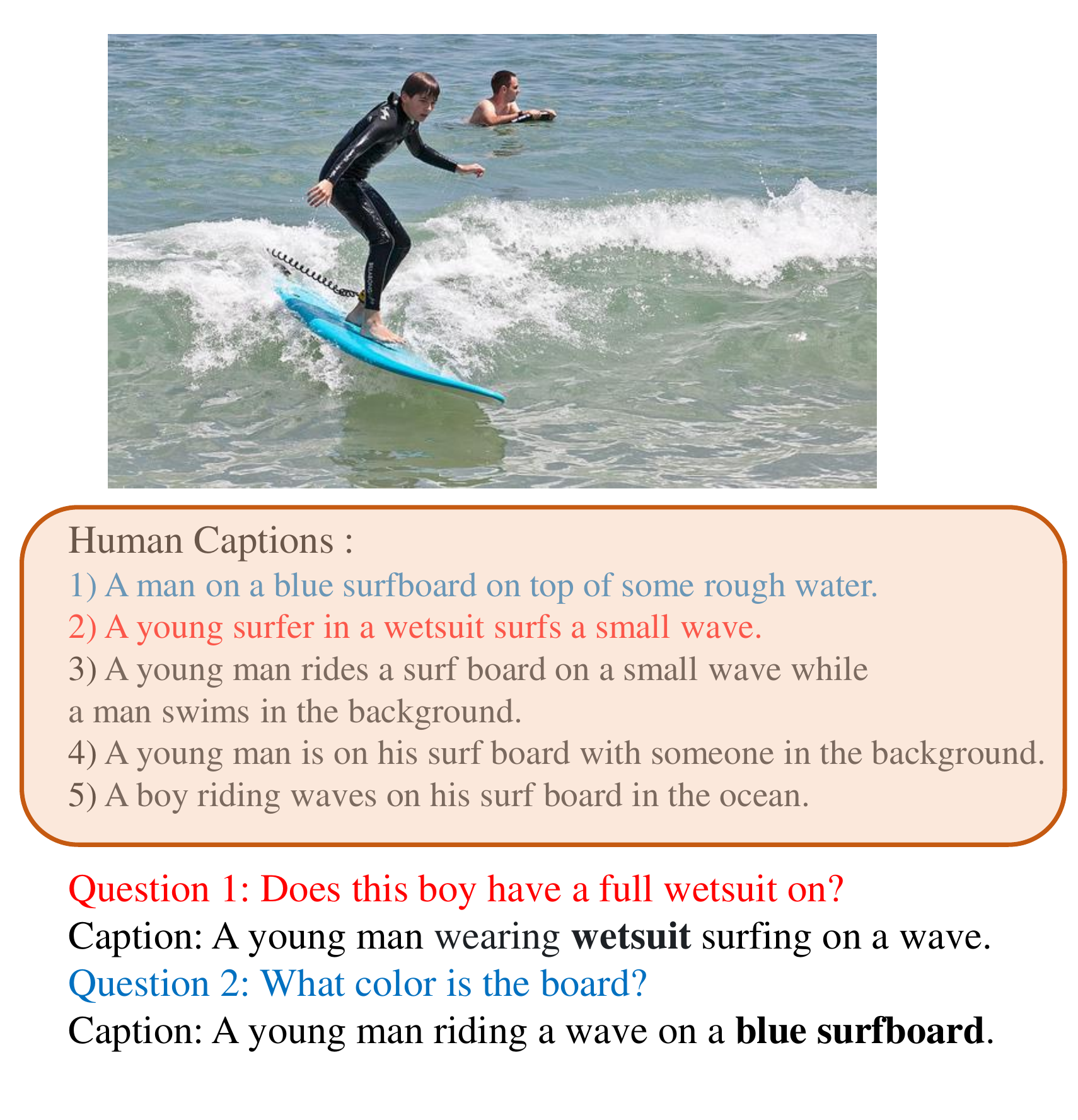}
\caption{Examples of our generated question-relevant captions. During the training phase, our model selects the most relevant human captions for each question (marked by the same color).}
\label{fig:relevance_caption_example}
\end{figure}

However,little VQA research works on exploiting textual features from the image which are able to tersely encode the necessary information to answer the questions. This information could be richer than the visual features in that the sentences have fewer structural constraints and can easily include the attributes of and relation among multiple objects. In fact, we observe that appropriate captions can be very useful for many VQA questions. In particular, we trained a model to answer visual questions for the VQA v2 challenge \cite{antol2015vqa} only using the human annotated captions \textbf{without} images and achieved a score of 59.6\%, outperforming a large number of VQA models that use image features. Existing work using captions for VQA has generated \textbf{question-agnostic} captions using a pretrained captioner \cite{li2018tell}. This approach can provide additional general information; however, this information is not guaranteed to be relevant to the given VQA question. 

We explore a novel approach that generates \textbf{question-relevant} image descriptions, which contain information that is directly relevant to a particular VQA question. Fig.\ \ref{fig:relevance_caption_example} shows examples of our generated captions given different questions. In order to encourage the generation of relevant captions, we propose a novel greedy algorithm that aims to minimize the cross entropy loss only for the most relevant and helpful gold-standard captions. Specifically, helpfulness is measured using the inner-product of the gradients from the caption generation loss and the VQA answer prediction loss. A positive inner-product means the two objective functions share some descent directions in the optimization process, and therefore indicates that the corresponding captions help the VQA training process.  

In order to incorporate the caption information, we propose a novel caption embedding module that, given the question and image features for a visual question, recognizes important words in the caption, and produces a caption embedding tailored for answer prediction. In addition, the caption embeddings are also utilized to adjust the visual top-down attention weights for each object.

Furthermore, generating question-relevant captions ensures that both image and question information is encoded in their joint representations, which reduces the risk of learning from question bias \cite{li2018tell} and ignoring the image content when high accuracy can be achieved from the questions alone.
 
Experimental evaluation of our approach shows significant improvements on VQA accuracy over our baseline \textbf{Up-Down} \cite{anderson2017bottom} model on the VQA v2 validation set \cite{antol2015vqa}, from $63.2\%$ to $67.1\%$ with gold-standard human captions from the COCO dataset \cite{chen2015microsoft}  and $65.8\%$ with automatically generated question-relevant captions. Our single model is able to score $68.4\%$ on the test-standard split, and an ensemble of $10$ models scores $69.7\%$.

\section{Background Related Work}
\subsection{Visual Question Answering} 
Recently, a large amount of attention-based deep-learning methods have been proposed for VQA, including top-down \cite{ren2015exploring,fukui2016multimodal,wu2016action,goyal2017making,li2018tell} and bottom-up attention methods \cite{anderson2017bottom,li2018vqa,wu2018self}. Specifically, a typical model first extracts image features using a pre-trained CNN, and then trains an RNN to encode the question, using an attention mechanism to focus on specific features of the image. Finally, both question and attended image features are used to predict the final answer.

However, answering visual questions requires not only information about the visual content but also common knowledge, which is usually too hard to directly learn from only a limited number of images with human annotated answers as supervision. However, comparatively little previous VQA research has worked on enriching the knowledge base. We are aware of two related papers. \newcite{li2018tell} use a pre-trained captioner to generate general captions and attributes with a fixed annotator and then use them to predict answers. Therefore, the captions they generate are not necessarily relevant to the question, and they may ignore image features needed for answer prediction. \newcite{narasimhan2018out} employed an out-of-the-box knowledge base and trained their model to filter out irrelevant facts. After that, graph convolutional networks use this knowledge to build connections to the relevant facts and predict the final answer. Unlike them, we generate captions to provide information that is directly relevant to the VQA process. 
\subsection{Image Captioning}
Most recent image captioning models are also attention-based deep-learning models  \cite{donahue2015long,karpathy2015deep,vinyals2015show,Luo_2018_CVPR,liu2018show}. With the help of large image description datasets \cite{chen2015microsoft}, these models have demonstrated remarkable results. Most of them encode the image using a CNN, and build an attentional RNN ($i.e.$ GRU \cite{cho2014learning}, LSTM \cite{hochreiter1997long}) on top of the image features as a language model to generate image captions. 

However, deep neural models still tend to generate general captions based on the most significant objects \cite{vijayakumar2016diverse}. Although previous works \cite{Luo_2018_CVPR,liu2018show} build captioning models that are encouraged to generate different captions with discriminability objectives, the captions are usually less informative and fail to describe most of the objects and their relationships diversely. In this work, we develop an approach to generating captions that directly focus on the critical objects in the VQA process and provide information that can help the VQA module predict the answer.
\begin{figure*}[!t]
\centering
\includegraphics[width=0.95\linewidth,trim={2cm 6.3cm 8.4cm 4.4cm},clip]{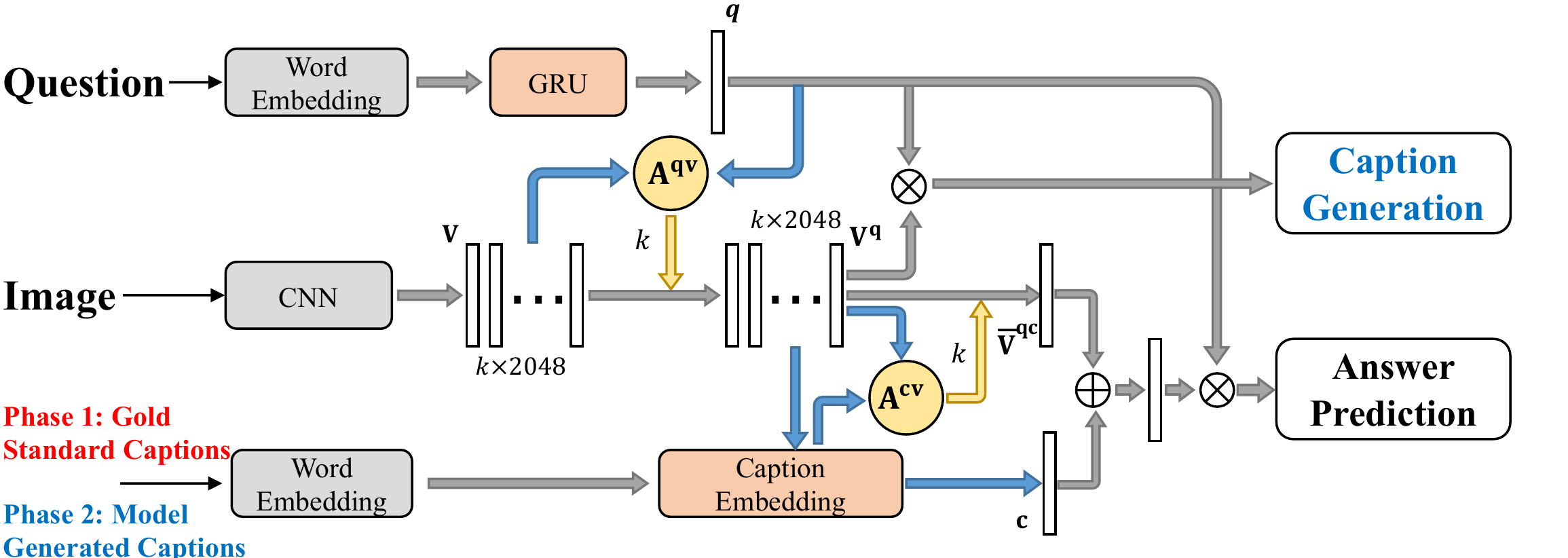}
\caption{Overall structure of our model that generates question-relevant captions to aid VQA. Our model is first trained to generate question-relevant captions as determined in an online fashion in phase 1. Then, the VQA model is fine-tuned with generated captions from the first phase to predict answers. 
$\otimes$ denotes element-wise multiplication and $\oplus$ denotes element-wise addition. Blue arrows denote fully-connected layers ($fc$) and yellow arrows denote attention embedding.}
\label{fig:overall_structure}
\end{figure*}
\section{Approach}
We first describe the overall structure of our joint model in Sec.\
\ref{sec:overview} and explain the foundational feature representations ($i.e.$ image, question and caption embeddings) in Sec.\ \ref{sec:feat_repr}. Then, the VQA module is presented in Sec.\ \ref{sec:vqa}, which takes advantage of the generated image captions to improve the VQA performance. In Sec.\ \ref{sec:ic}, we explain the image captioning module which generates question-relevant captions. Finally, the training and implementation details are provided in Sec.\ \ref{sec:training}.

\subsection{Overview}
\label{sec:overview}
As illustrated in Fig.\ \ref{fig:overall_structure}, the proposed model first extracts image features $\textbf{V}=\{\textbf{v}_1, \textbf{v}_2, ..., \textbf{v}_K\}$ using bottom-up attention and question features $\textbf{q}$ to produce their joint representation and then generates question-related captions. 
Next, our caption embedding module encodes the generated captions as caption features $\textbf{c}$ as detailed in Sec.\ \ref{sec:feat_repr}. 
After that, both question features $\textbf{q}$ and caption features $\textbf{c}$ are utilized to generate the visual attention $\textbf{A}^{cv}$ to weight the images' feature set $\textbf{V}$, producing attended image features $\overline{\textbf{v}}^{qc}$. Finally, we add $\overline{\textbf{v}}^{qc}$ to the caption features $\textbf{c}$ and further perform element-wise multiplication with the question features $\textbf{q}$ \cite{anderson2017bottom} to produce the joint representation of the question, image and caption, which is then used to predict the answer.

\subsection{Feature Representation}
\label{sec:feat_repr}
In this section, we explain the details of this joint representation. We use $f(x)$ to denote fully-connected layers, where $f(x)$ $ = \text{LReLU}(Wx + b)$ with input features $x$ and ignore the notation of weights and biases for simplicity, where these $fc$ layers do not share weights. $\text{LReLU}$ denotes a Leaky ReLU \cite{he2015delving}.\\

\noindent\textbf{Image and Question Embedding}\\
We use object detection as bottom-up attention \cite{anderson2017bottom}, which provides salient image regions with clear boundaries. In particular, we use a Faster R-CNN head \cite{girshick2015fast} in conjunction with a ResNet-101 base network \cite{he2016deep} as our detection module. The detection head is first pre-trained on the Visual Genome dataset \cite{krishna2017visual} and is capable of detecting $1,600$ objects categories and $400$ attributes. To generate an output set of image features $\textbf{V}$, we take the final detection outputs and perform non-maximum suppression (NMS) for each object category using an IoU threshold of $0.7$. Finally, a fixed number of 36 detected objects for each image are extracted as the image features (a $2,048$ dimensional vector for each object) as suggested by \newcite{teney2017tips}.

For the question embedding, we use a standard GRU \cite{cho2014learning} with $1,280$ hidden units and extract the output of the hidden units at the final time step as the question features $\textbf{q}$.
Following \newcite{anderson2017bottom}, the question features $\textbf{q}$ and image feature set $\textbf{V}$ are further embedded together to produce a question-attended image feature set $\textbf{V}^q$ via  question visual-attention $\textbf{A}^{qv}$ as illustrated in Fig. \ref{fig:overall_structure}.\\ 

\noindent\textbf{Caption Embedding}\\
Our novel caption embedding module takes as input the question-attended image feature set $\textbf{V}^q$, question features $\textbf{q}$, and  $C$ captions $\textbf{W}^c_i = \{w^c_{i, 1}, w^c_{i, 2}, ..., w^c_{i,T}\}$, where $T$ denotes the length of the captions and $i=1,...,C$ are the caption indices, and then produces the caption features $\textbf{c}$. 

\begin{figure}[h]
\centering
\includegraphics[width=0.8\linewidth,trim={0cm 8.5cm 22.5cm 0.5cm},clip]{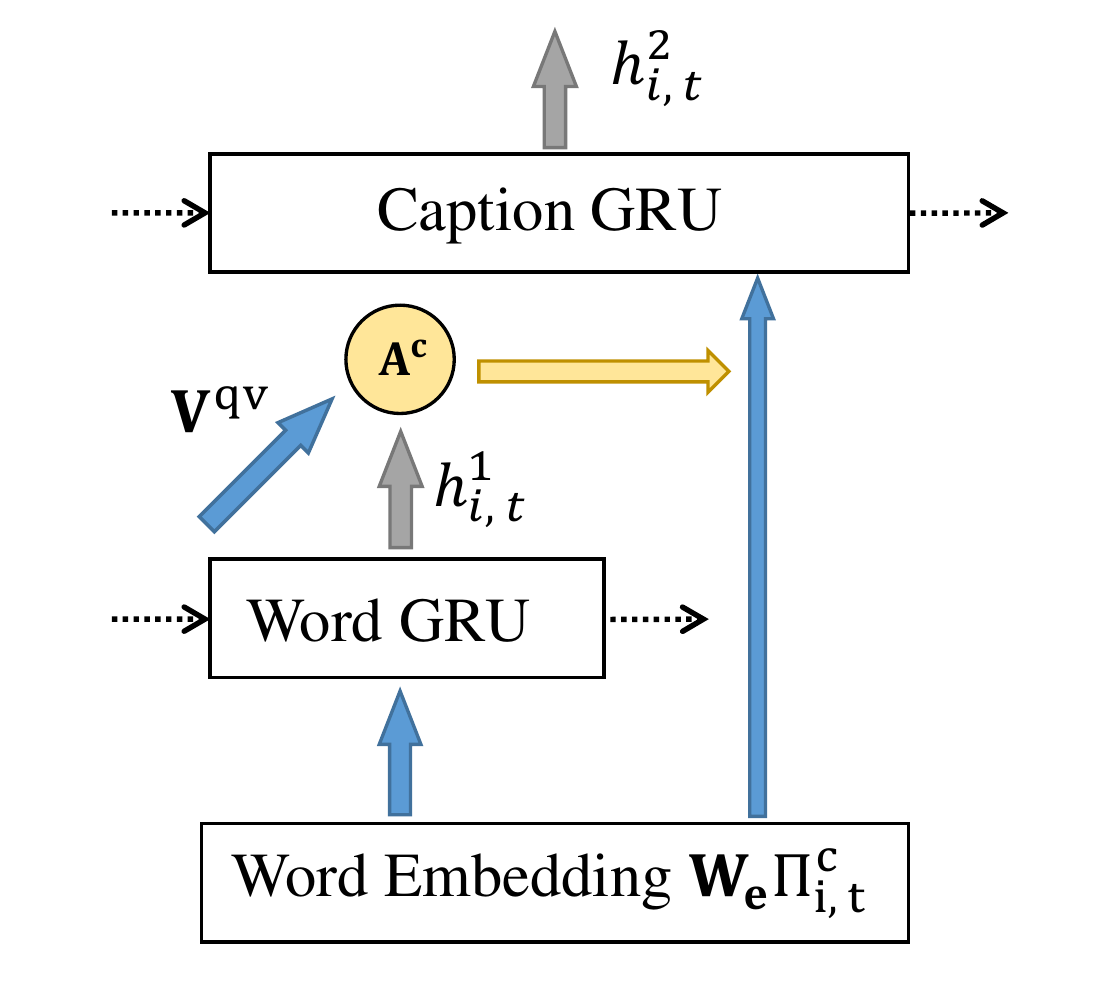}
\caption{Overview of the caption embedding module. The Word GRU is used to generate attention to identify the relevant words in each caption, and the Caption GRU generates the final caption embedding. We use question-attended image features $\textbf{V}^{qv}$ to compute the attention. Blue arrows denote $fc$ layers and yellow arrows denote attention embedding.}
\label{fig:caption_module}
\end{figure}

The goals of the caption module are to serve as a knowledge supplement to aid VQA, and to provide additional clues to identify the relevant objects better and adjust the top-down attention weights.
To achieve this, as illustrated in Fig.\ \ref{fig:caption_module}, we use a two-layer GRU architecture. The first-layer GRU (called the Word GRU)  sequentially encodes the words in a caption $\textbf{W}^c_i$ at each time step as $h^1_{i,t}$. 
\begin{align}
    h^1_{i,t} &= \text{GRU}( \textbf{W}_e \Pi^c_{i,t},~ h^1_{i, t-1}) 
    \label{eq:caption_word_attention}
\end{align}   
where $\textbf{W}_e$ is the word embedding matrix, and $\Pi^c_{i,t}$ is the one-hot embedding for the word $w^c_{i,t}$.

Then, we design a caption attention module $\textbf{A}^c$ which utilizes the question-attended feature set $\textbf{V}^{q}$, question features $\textbf{q}$, and $h^1_{i,t}$ to generate the attention weight on the current word in order to indicate its importance. Specifically, the Word GRU first encodes the words embedding $\Pi^c_{i,t}$ in Eq. \ref{eq:caption_word_attention}, and then we feed the outputs $h^1_{i,t}$ and $\textbf{V}^{q}$ to the attention module $\textbf{A}^c$ as shown in Eq. \ref{eq:caption_attention}. 
\begin{align}
    \overline{\textbf{v}}^{q} &= \sum_{k = 1}^{K} \textbf{v}_k^{q}\\
    a^{c}_{i,t} &= h^1_{i,t} \circ f(\overline{\textbf{v}}^{q}) + h^1_{i,t} \circ f(\textbf{q}) \\
    \label{eq:caption_attention}
    \alpha^c_{i, t} &= \sigma(a^c_{i, t})
\end{align}
where $\sigma$ denotes the sigmoid function, and $K$ is the number of objects in the bottom-up attention.

Next, the attended words in the caption are used to produce the final caption representation in Eq. \ref{eq:attended_caption_feed_foreward} via the Caption GRU. Since the goal is to gather more information, we perform element-wise max pooling across the representations of all of the input captions $\textbf{c}_i$ in Eq.\ \ref{eq:caption:maxp}.
\begin{align} 
\label{eq:attended_caption_feed_foreward}
    h^2_{i,t} &= \text{GRU}( \alpha^c_{i, t}\textbf{W}_e \Pi^c_{i,t}, ~ h^2_{i,t-1})\\
    \textbf{c}_i &= f(h^2_{i,T})\\
    \label{eq:caption:maxp}
    \textbf{c} &= max(\textbf{c}_i)
\end{align}
where $max$ denotes the element-wise max pooling across all of caption representations $\textbf{c}_i$ of the image.

\subsection{VQA Module}
\label{sec:vqa}
This section describes the details of the VQA module. The generated captions are usually capable of capturing relations among the question-relevant objects; however these relations are absent in the bottom-up attention. Therefore, our VQA module utilizes the caption embeddings $\textbf{c}$ to adjust the top-down attention weights in VQA in order to produce the final caption-attended features $\overline{\textbf{v}}^{qc}$ in Eq.\ \ref{eq:question-caption-attended}:
\begin{align}
    a^{cv}_{k} &= f(f(\textbf{c})\circ f(\overline{\textbf{v}}^q_k))\\
    \label{eq:image_caption_attention}
    \alpha^{cv}_{k} &=  softmax(a^{cv}_{c,k})\\
    \label{eq:question-caption-attended}
    \overline{\textbf{v}}^{qc} &=  \sum_{k}^{K}  \textbf{v}^{q}_{k} \alpha^{cv}_{k}
\end{align}
where $k$ traverses the $K$ objects features.\\
To better incorporate the information from the captions into the VQA process, we add the caption features $\textbf{c}$ to the attended image features $\overline{\textbf{v}}^{qc}$, and then element-wise multiply by the question features as shown in Eq. \ref{eq:vqa_repr}:
\begin{align}
\label{eq:vqa_repr}
    \textbf{h} &= \textbf{q} \circ (f(\overline{\textbf{v}}^{qc}) + f(\textbf{c}))\\
    \hat{s} &= \sigma(f(\textbf{h}))
\end{align}

We frame the answer prediction task as a multi-label regression problem \cite{anderson2017bottom}. In particular, we use the soft scores in the gold-standard VQA-v2 data (which are used in the evaluation metric), as labels to supervise the sigmoid-normalized predictions as shown in Eq. \ref{eq:vqa_loss}:
\begin{equation}
\label{eq:vqa_loss}
    \mathcal{L}^{vqa} = -\sum^N_{j=1} s_{j}\log\hat{s}_{j} + (1-s_{j})\log(1-\hat{s}_{j})
\end{equation}
where the index $j$ runs  over 
$N$ candidate answers and $s$ are the soft answer scores.

In case of multiple feasible answers, the soft scores capture the occasional uncertainty in the ground-truth annotations. As suggested by \newcite{teney2017tips}, we collect the candidate answers that appear more than $8$ times in the training set, which results in $3,129$ answer candidates. 

\subsection{Image Captioning Module}
\label{sec:ic}
We adopt an image captioning module similar to that of \newcite{anderson2017bottom}, which takes the object detection features as inputs and learns attention weights over those objects' features in order to predict the next word at each step. The key difference between our module and theirs lies in the input features and the caption supervision. Specifically, we use the question-attended image features $\textbf{V}^q$ as inputs, and only use the most relevant caption, which is automatically determined in an online fashion (detailed below), for each question-image pair to train the captioning module. This ensures that only question-relevant captions are generated.\\

\noindent\textbf{Selecting Relevant Captions for Training}\\
 Previously, \newcite{li2018vqa} selected relevant captions for VQA based on word similarities between captions and questions, however, their approach does not take into account the details of the VQA process. In contrast, during training, our approach dynamically determines for each problem, the caption that will most improve VQA.  
We do this by updating with a shared descent direction \cite{wu2018dynamic} which decreases the loss for {\it both} captioning and VQA. This ensures a consistent target for both the image captioning module and the VQA module in the optimization process.

During training, we compute the cross-entropy loss for the $i$-th caption using Eq.\ \ref{eq:caption_loss}, and back-propagate the gradients only from the most relevant caption determined by solving Eq.\ \ref{pro:caption_selection}. 
\begin{align}
    \mathcal{L}^c_{i} =& -\sum^T_{t=1}\log(p(w^{c}_{i,t}|w^c_{i,t-1}))
    \label{eq:caption_loss}
    \end{align}
In particular, we require the inner product of the current gradient vectors from the predicted answer and the human captions to be greater than a positive constant $\xi$, and further select the caption that maximizes that inner product. 
\begin{align}
\begin{split}
    \underset{i}{\arg \max}& \sum_{k=0}^{K}  \left( \frac{\partial \hat{s}_{\text{pred}}}{\partial \textbf{v}^{q}_k}\right) ^T\frac{\partial \log (p(\textbf{W}^{c}_i))}{\partial \textbf{v}^{q}_k} \\
s.t.~\sum_{k=0}^{K} & \left( \frac{\partial \hat{s}_{\text{pred}} }{\partial \textbf{v}^{q}_k}\right) ^T\frac{\partial \log (p(\textbf{W}^{c}_i))}{\partial \textbf{v}^{q}_k} > \xi 
    \label{pro:caption_selection}
\end{split}
\end{align}
where the $\hat{s}_{\text{pred}}$ is the logit\footnote{The input to the softmax function.} for the predicted answer, $\textbf{W}^{c}_i$ denotes the $i$-th human caption for the image and $k$ traverses the $K$ object features.

Therefore, given the solution to Eq.\ \ref{pro:caption_selection}, $i^{\star}$, the final loss of our joint model is the sum of the VQA loss and the captioning loss for the selected captions as shown in Eq.\ \ref{eq:total_loss}. If Eq.\ \ref{pro:caption_selection} has no feasible solution, we ignore the caption loss. \\
\begin{equation}
    \mathcal{L} = \mathcal{L}^{vqa} + \mathcal{L}^c_{i^{\star}}
    \label{eq:total_loss}
\end{equation}

\subsection{Training and Implementation Details}
\label{sec:training}
We train our joint model using the AdaMax optimizer \cite{kingma2014adam} with a batch size of $384$ and a learning rate of 0.002 as suggested by \newcite{teney2017tips}. We use the validation set for VQA v2 to tune the initial learning rate and the number of epochs, yielding the highest overall VQA score. We use $1,280$ hidden units in the question embedding and attention model in the VQA module with $36$ object detection features for each image. For captioning models, the dimension of the LSTM hidden state, image feature embedding, and word embedding are all set to 512. We also use Glove vectors \cite{pennington2014glove} to initialize the word embedding matrix in the caption embedding module.

 We initialize the training process with human annotated captions from the COCO dataset \cite{chen2015microsoft} and pre-train the VQA and  caption-generation modules for 20 epochs with the final joint loss in Eq.\ \ref{eq:total_loss}.  After that, we generate question-relevant captions for all question-image pairs in the COCO train, validation, and test sets. In particular, we sample $5$ captions per question-image pair. We fine-tune our model using the generated captions with 0.25 $\times$ learning-rate for another 10 epochs.

\begin{table*}[!t]
\centering
\begin{tabular}{l|ccc|c}
\hline \toprule
                   & \multicolumn{4}{c}{Test-standard} \\\hline
                       & Yes/No  & Num   & Other  &All\\ \hline\hline
Prior \cite{goyal2017making} &61.20 & 0.36 &1.17 & 25.98 \\
Language-only \cite{goyal2017making} & 67.01 &31.55 & 27.37 & 44.26\\
MCB \cite{fukui2016multimodal} & 78.82 & 38.28 & 53.36 & 62.27 \\
Up-Down \cite{anderson2017bottom}   & 82.20    & 43.90  & 56.26 & 65.32   \\
VQA-E \cite{li2018vqa}   & 83.22 & 43.58 & 56.79 & 66.31   \\
\textbf{Ours}(single) & \textbf{84.69}   &  \textbf{46.75}  &    \textbf{59.30}     & \textbf{68.37}    \\\hline
\textbf{Ours}(Ensemble-10)  &\textbf{86.15}   &  \textbf{47.41} &  \textbf{60.41}   & \textbf{69.66}    \\\bottomrule
\end{tabular}
\caption{Comparison of our results on VQA with the state-of-the-art methods on the test-standard data. Accuracies in percentage (\%) are reported.}
\label{tab:vqa_compare}
\end{table*}
\section{Experiments}
We perform extensive experiments and ablation studies to evaluate our joint model on VQA. 

\subsection{Datasets and Evaluation Metrics}
\noindent\textbf{VQA Dataset}\\
We use the VQA v2.0 dataset \cite{antol2015vqa} for the evaluation of our proposed joint model, where the answers are balanced in order to minimize the effectiveness of learning dataset priors. This dataset is used in the VQA 2018 challenge and contains over $1.1$M questions from the over $200$K images in the MSCOCO 2015 dataset \cite{chen2015microsoft}.

Following \newcite{anderson2017bottom}, we perform standard text pre-processing and tokenization. In particular, questions are first converted to lower case and then trimmed to a maximum of $14$ words, and the words that appear less than $5$ times are replaced with an ``$<$unk$>$'' token. To evaluate answer quality, we report accuracies using the official VQA metric using soft scores, which accounts for the occasional disagreement between annotators for the ground truth answers.\\

\noindent\textbf{Image Captioning Dataset}\\
We use the MSCOCO 2014 dataset \cite{chen2015microsoft} for the image caption module. To maintain consistency with the VQA tasks, we use the dataset's official configuration that includes $82,372$ images for training and $40,504$ for validation. 
Similar to the VQA question pre-processing, we first convert all sentences to lower case, tokenizing on white spaces, and filtering words that do not occur at least $5$ times. 

\subsection{Results on VQA}
We first report the experimental results on the VQA task and compare our results with the state-of-the-art methods in this section. After that, we perform ablation studies to verify the contribution of additional knowledge from the generated captions, and the effectiveness of using caption representations to adjust the top-down visual attention weights. 

As demonstrated in Table \ref{tab:vqa_compare}, our single model outperforms other state-of-the-art single models by a clear margin, $i.e.\ 2.06\%$, which indicates the effectiveness of including caption features as additional inputs. In particular, we observe that our single model outperforms other methods, especially in the 'Num' and 'Other' categories. This is because the generated captions are capable of providing more numerical clues for answering the 'Num' questions, since the captions can describe the number of relevant objects and provide general knowledge for answering the 'Other' questions. Furthermore, an ensemble of $10$ models with different initialization seeds results in a score of 69.7\% for the test-standard set.
\begin{figure*}[!t]
\centering
\includegraphics[width=\linewidth,trim={0cm 0.5cm 2cm 0cm},clip]{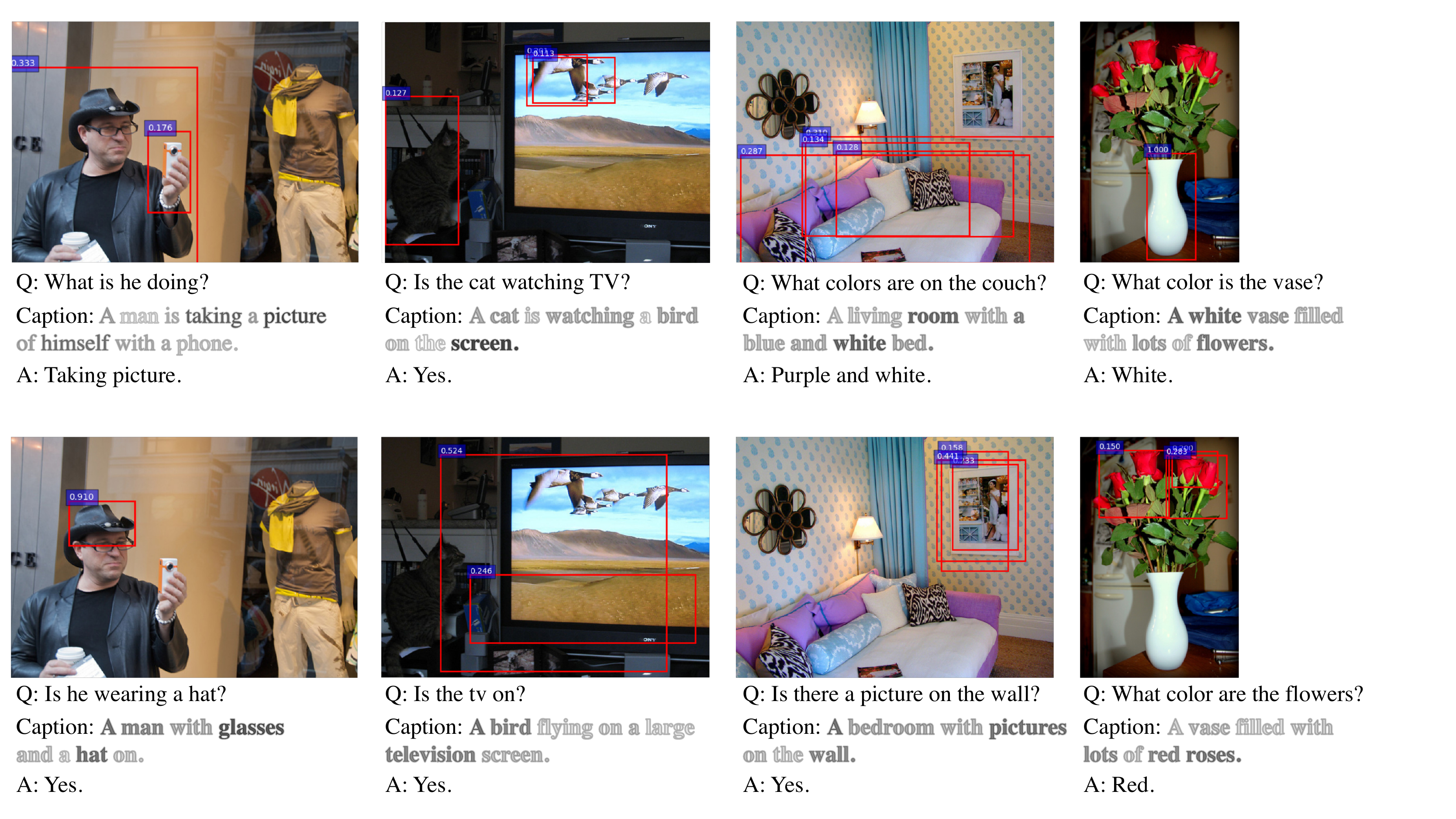}
\caption{Examples of our generated question-relevant captions. The influential objects with attention weights greater than 0.1 are indicated by bounding boxes (annotated with their visual attention weights in the blue box), and the gray-scale levels in the caption words indicate the word attentions from the caption embedding module.}
\label{fig:attention_analysis}
\end{figure*}

Fig.\ \ref{fig:attention_analysis} shows several examples of our generated question-relevant captions. These examples illustrate how different captions are generated for the same image when the question is changed. They also
show how the objects in the image that are important to answering the question are described in the question-relevant captions.\\

\noindent\textbf{Comparison Between Using Generated and Human Captions}\\
Next, we analyze the difference between using automatically generated captions and using those provided by human annotators. 
In particular, we train our model with generated question-agnostic captions using the Up-Down \cite{anderson2017bottom} captioner, question-relevant captions from our caption generation module, and human annotated captions from the COCO dataset. 

\begin{table}[h]
\centering
\begin{tabular}{l|c}
\hline \toprule
                  & Validation \\ \hline
Up-Down \cite{anderson2017bottom} &  63.2\\
Ours with Up-Down captions &    64.6\\
Ours with our generated captions &    65.8\\
Ours with human captions &    \textbf{67.1} \\\bottomrule
\end{tabular}
\caption{Comparison of the performance using generated and human captions. Both of them provide significant improvements to the baseline model. However, there is still a reasonable gap between generated and human captions.}
\label{tab:gen_anno_compare}
\end{table}

As demonstrated in Table \ref{tab:gen_anno_compare}, our model gains about 4\% improvement from using human captions and 2.5\% improvement from our generated question-relevant captions on the validation set. This indicates the insufficiency of directly answering visual questions using a limited number of detection features, and the utility of incorporating additional information about the images. We also observe that our generated question-relevant captions trained with our caption selection strategy provide more helpful clues for the VQA process than the question-agnostic Up-Down captions, outperforming their captions by 1.2\%. \\

\begin{figure}[!h]
\centering
\includegraphics[width=\linewidth,trim={1cm 7cm 13cm 1cm},clip]{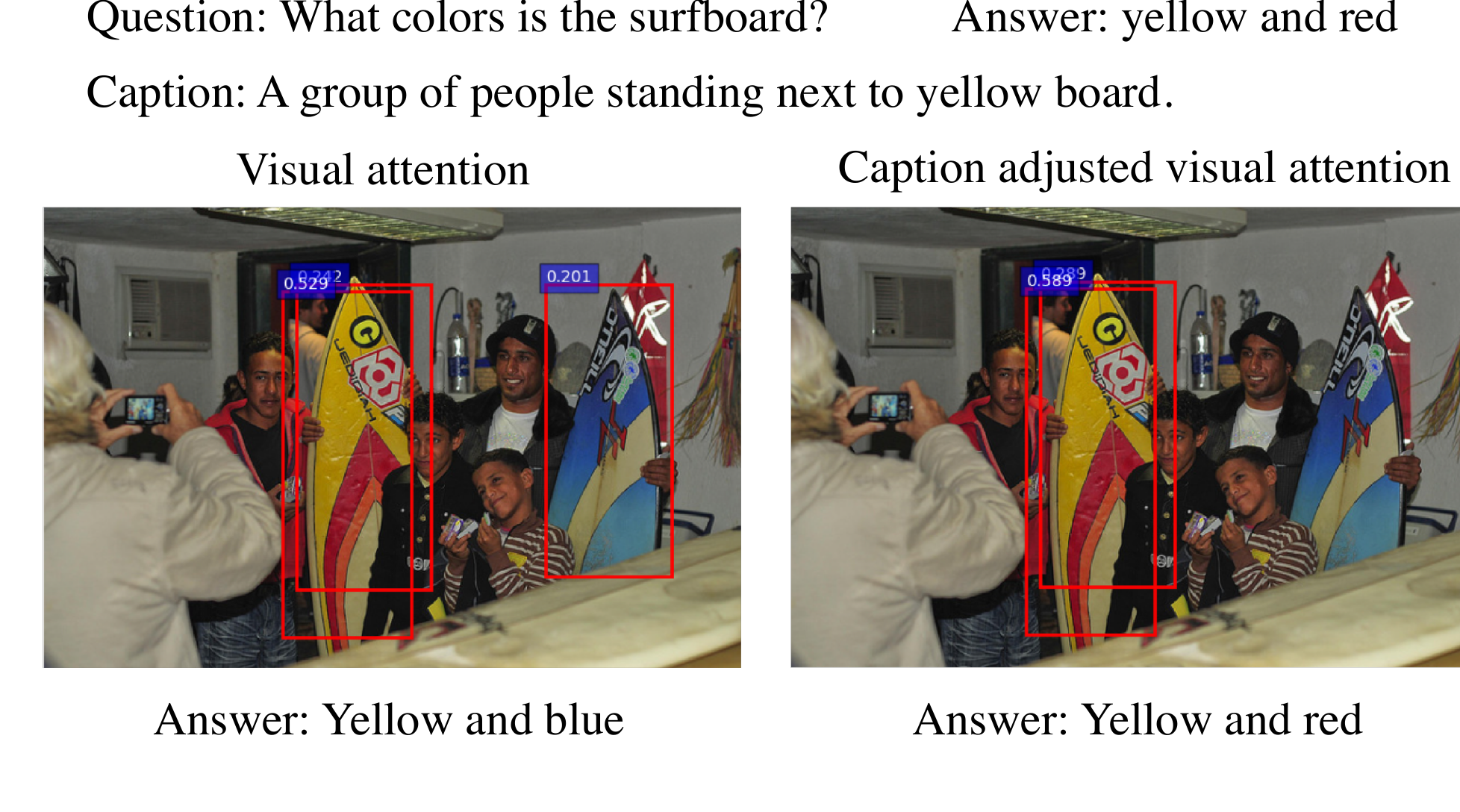}
\caption{An example of caption attention adjustment. The question-relevant caption helps the VQA module adjust the visual attention from both the yellow board and the blue sail to the yellow board only. }
\label{fig:caption_adjust}
\end{figure}

\noindent\textbf{Effectiveness of Adjusting Top-Down Attention}\\
In this section, we quantitatively analyze the effectiveness of utilizing captions to adjust the top-down attention weights, in addition to the advantage of providing additional information.
In particular, we compare our model with a baseline version where the top-down attention-weight adjustment factor $\textbf{A}^{cv}$ is manually set to 1.0 (resulting in no adjustment). 

As demonstrated in Tables \ref{tab:relationship} and \ref{tab:relationship1}, we observe an improvement when using caption features to adjust the attention weights. This indicates that the caption features help the model to more robustly locate the objects that are helpful to the VQA process. We use $w~\mathcal{CAA}$ to indicate with caption attention adjustment and $w/o~\mathcal{CAA}$ to indicate without it. Fig.\ \ref{fig:caption_adjust} illustrates an example of caption attention adjustment. Without $\mathcal{CAA}$, the top-down visual attention focuses on both the yellow surfboard and the blue sail, generating the incorrect answer ``yellow and blue.''. However, with ``yellow board'' in the caption, the caption attention adjustment ($\mathcal{CAA}$) helps the VQA module focus attention just on the yellow surfboard, thereby generating the correct answer ``yellow and red'' (since there is some red coloring in the surfboard).
 \begin{table}[h]
\centering
\begin{tabular}{l|cccc}
\hline \toprule
                  & \multicolumn{4}{c}{Test-standard} \\\hline
                  & {\footnotesize All }   & {\footnotesize Yes/No}  & {\footnotesize Num}   & {\footnotesize Other}  \\ \hline\hline
Up-Down              & 65.3   & 82.2    & 43.9  & 56.3  \\
Ours $w/o~\mathcal{CAA}$ & 67.4   & 84.0 & 44.5 & 57.9\\
Ours $w~\mathcal{CAA}$ & 68.4  &  84.7 & 46.8 &  59.3     \\\bottomrule
\end{tabular}
\caption{Evaluation of the effectiveness of caption-based attention adjustment ($\mathcal{CAA}$) on the test-standard data. Accuracies in percentage (\%) are reported.}
\label{tab:relationship}
\end{table}

\begin{table}[h]
\centering
\begin{tabular}{l|cccc}
\hline\toprule
                  & \multicolumn{4}{c}{Validation} \\  \hline
                  & {\footnotesize All }   & {\footnotesize Yes/No}  & {\footnotesize Num}   & {\footnotesize Other}  \\ \hline\hline
Up-Down  & 63.2  & 80.3    & 42.8 & 55.8   \\
Ours $w/o~\mathcal{CAA}$  & 65.2  &  82.1 & 43.6 & 55.8\\
Ours $w~\mathcal{CAA}$ &  65.8   &  82.6&  43.9& 56.4\\\bottomrule
\end{tabular}
\caption{Evaluation of the effectiveness of $\mathcal{CAA}$ on the validation data. Accuracies in percentage (\%) are reported.}
\label{tab:relationship1}
\end{table}

Next, in order to directly demonstrate that our generated question-relevant captions help the model to focus on more relevant objects via attention adjustment, we compare the differences between the generated visual attention and human-annotated important objects from the VQA-X dataset \cite{park2018multimodal}, which has been used to train and evaluate multimodal (visual and textual) VQA explanation \cite{wu2018faithful}. The VQA-X dataset contains $2,000$ question-image pairs from the VQA v2 validation set with human annotations indicating the objects which most influence the answer to the question. In particular, we used Earth Mover Distance (EMD) \cite{rubner2000earth} to compare the highly-attended objects in the VQA process to the objects highlighted by human judges.
This style of evaluation using EMD has previously been employed to compare automatic visual explanations to human-attention annotations \cite{selvaraju2017grad,park2018multimodal}.

We resize all of the $2,000$ human annotations in VQA-X dataset to 14$\times$14 and adjust the object bounding boxes in the images accordingly. Next, we assign the top-down attention weights to the corresponding bounding boxes, both before and after caption attention adjustment, and add up the weights of all $36$ detections. Then, we normalize attention weights over the 14 $\times$ 14 resized images to sum to one, and finally compute the EMD between the normalized visual attentions and the human annotations.

Table \ref{tab:emd} reports the EMD results for the attentions weights both before and after the caption attention adjustments. 
\begin{table}[!h]
\centering
\begin{tabular}{l|c|c|c}
\hline \toprule
     &  $w/o~\mathcal{CAA}$  & $w~\mathcal{CAA}$ & Human  \\\hline
EMD  & 2.38  & 2.30    & 2.25    \\\bottomrule
\end{tabular}
\caption{EMD results comparing the top-down attention weights (with or without caption attention adjustments) to human attention-annotation from the VQA-X dataset. Results are shown for both automatically generated captions and human captions. Lower EMD indicates a closer match to human attention.}
\label{tab:emd}
\end{table}

The results indicate that caption attention adjustment improves the match between automated attention and human-annotated attention, even though the approach is not trained on supervised data for human attention. Not surprisingly, human captions provide a bit more improvement than automatically generated ones.

\section{Conclusion}
In this work, we have explored how generating question-relevant image captions can improve VQA performance.  In particular, we present a model which jointly generates question-related captions  and uses them to provide additional information to aid VQA. This approach only utilizes existing image-caption datasets, automatically determining which captions are relevant to a given question. In particular, we design the training algorithm to only update the network parameters in the optimization process when the  caption generation and VQA tasks agree on the direction of change. Our single model joint system outperforms the current state-of-the-art single model for VQA. 
\section*{Acknowledgement}
This research was supported by the DARPA XAI program under a grant from AFRL. 
\bibliography{acl2019}
\bibliographystyle{acl_natbib}
\end{document}